# DEEP LEARNING-BASED SEMANTIC SEGMENTATION OF DENTAL RADIOGRAPHS: COMPARATIVE EVALUATION OF PSPNET, DEEPLABV3, UNET, AND UNET++

Malithi Nawarathne

*Faculty of Computing, Sri Lanka Institute of Information Technology, Sri Lanka*

https://orcid.org/0009-0002-2489-2421

Nadeesha Herath Mudiyanselage

*Department of Computer Science, Faculty of Science, University of Helsinki, Finland*

https://orcid.org/0000-0002-4708-5561

Lihini Sangeetha

*Computing Centre, Faculty of Engineering, University of Peradeniya, Sri Lanka*

https://orcid.org/0009-0001-1410-7840

## ABSTRACT

*Accurate and efficient analysis of dental radiographs has become increasingly important for disease diagnosis and clinical decision-making. In particular, dental radiographs segmentation plays a critical role as it is challenging due to several reasons, such as low contrast, noise in the images, variations in the image quality, etc. Previous studies have primarily addressed these challenges using convolutional neural networks (CNN)-based models. However, direct comparison of these segmentation models is still challenging due to the variations among the experimental settings. Therefore, we mainly compared four segmentation models, including Pyramid Scene Parsing Network (PSPNet), DeepLabV3, UNet, and UNet++, through a fair, controlled experimentation setting under varying conditions such as image sizes and batch sizes. For this, we used 1269 dental radiography images, and prior to training, we implemented a series of preprocessing techniques. To assess the models, we used several metrics, including Dice coefficient (DICE), Intersection over Union (IoU), accuracy, precision, training and validation losses. According to our results, UNet++ demonstrated the highest IoU score of 0.6719, and a DICE value of 0.7869, showing superior results in terms of segmentation, specifically when configured with an image size of 300×300 and a batch size of 16. In addition, we also explored the effect of image resolutions and batch sizes on model performance and accuracy under different architectures. Therefore, this study provides insights into the strengths of different segmentation models and the potential of deep learning–based semantic segmentation methods for improving automated dental radiograph analysis*.

Corresponding Author: Malithi Nawarathne email: malithi.n@sliit.lk

# 1. INTRODUCTION

Teeth segmentation and orientation are fundamental in modern oral healthcare, enabling precise diagnosis, treatment planning, and dental implant design (Dhar *et al.*, 2023). Dental radiography serves as a primary medium for diagnosis, treatment planning, and the longitudinal monitoring of oral health based applications (Rad *et al.*, 2013; Brahmi, Jdey and Drira, 2024). For dental practitioners, the ability to separate individual teeth and structures is essential for automatic dental chart filing, which reduces administrative burdens, enables the identification of specific pathologies or anatomical anomalies with high precision, and creates a structured dataset for further diagnostic analysis.

In the past, dental structures were identified and labelled manually which is a time-consuming process that is prone to inter-observer variability. To overcome these inefficiencies researchers have shifted towards automated segmentation frameworks (Brahmi, Jdey and Drira, 2024; Zhou *et al.*, 2025). Recent advances in deep learning (DL), especially convolutional neural networks and UNet architectures, have set a new standard for accuracy. DL frameworks are capable of processing dental images with the high complexity and noise that traditional methods cannot handle, offering high-precision tooth instance segmentation (Fındık *et al.*, 2025) and transparent, verifiable results to the clinicians. This also enables the creation of diagnostic reports by automatically filling in dental charts (Muramatsu *et al.*, 2021) transforming pixels directly into clinical documentation. Thus, it is evident that automated segmentation methods, particularly those based on deep learning and artificial intelligence, offer several transformative clinical advantages for contemporary dental diagnostic workflows. Instead of manual identification, dental practitioners can shift to automated identification with high precision, which helps them provide more accurate diagnosis and manage their administrative activities. Therefore, the present study aimed to explore and compare four deep learning-based semantic segmentation models using dental radiographs.

This paper is structured as follows: In Section II we discuss the previous literature related to the topic of this study. We also describe the methodology that we followed in Section III while we explain the results that we obtained in detail in Section IV. Finally, we conclude the study in Section V.

# 2. RELATED WORK

Dental image analysis has evolved significantly with advances in image segmentation, classification, and explainability using machine learning and deep learning techniques. This section reviews key methodologies and contributions in tooth segmentation, classification, and the integration of explainable artificial intelligence (XAI).

Early research in dental radiography segmentation has explored traditional image processing methods. For instance, Rad et al. (2013) discussed existing methods of segmentation that have been applied to dental X-ray images, including thresholding and integral projection, active contour without edge, region growing, and level-set methods (Rad *et al.*, 2013).

With the rise of deep learning, more robust and automated solutions have emerged. Muramatsu et al. (2020) investigated the feasibility of automatic filing of the dental chart using CNNs. The proposed method plays an important role in the automatic filing of dental charts for forensic identification and the preprocessing of dental images for prescreening purposes. In addition, the comparison of single and multi-sized input models in their paper highlighted the importance of neighboring information in tooth type classification.

Furthermore, several studies have improved dental radiograph segmentation using deep learning techniques achieving a higher accuracy. Furthermore, Datta (2023) proposed a technique that automatically segments the teeth regions and assists in further analysis on both peri-apical and panoramic types of dental radiographic images (Datta, Chaki and Modak, 2023). For this, they used neutrosophic logic along with the fuzzy c-means algorithm. Similarly, (Brahmi, Jdey and Drira, 2024) showed that the DL models facilitate dentists to perform precise tooth identification for the categorization of dental pathologies, assessment of prosthetic structures, and reduction of errors .

Moreover, few studies have developed some advanced segmentation models to enhance the dental structures detection. For example, Deb, Deb and Dhar (2025) used an EfficientNet-based encoder and a decoder with grid-based attention gates and a parallel spatial and channel squeeze-and-excitation (P-scSE) module to develop an end-to-end instance segmentation network. They have also introduced oriented bounding boxes (OBB) generation using principal component analysis (PCA) to accurately determine the orientation and spatial boundaries of dental structures.

The authors Fındık et al. (2025) suggested a GAN-based model, which is trained to learn the relationships in the space between the original panoramic radiography images and the ground truth images. The latter include the boundaries of image components. This algorithm was tested on the UESB dataset. In order to make comparisons, the performance of the algorithm is compared with the one of UNet and state-of-the-art methods tested on the same dataset.

Some approaches have combined segmentation with higher-level tasks. A two-stage pipeline model comprising a segmentation module and a classification module was designed by Zhou et al. (2025). The segmentation module integrates the 3D TransUNet model and fine-tuned nnUNetv2 framework to predict tooth location and numbering.

Similarly, a few recent studies have emphasized the transparency and clinical interpretability through explainable artificial intelligence. Sreeram, R. and MN (2025) used XAI techniques, particularly Gradient-weighted Class Activation Mapping (Grad-CAM), to generate heat maps that highlight significant areas within the radiographs, thereby enhancing the model's transparency. They proposed a hybrid strategy that integrates attention-based segmentation with both supervised and self-supervised learning models. The hybrid model combining U-Net and Grad-CAM achieved an Intersection over Union (IoU) score of 85.5, a Dice coefficient of 92.7, an accuracy of 98.2%, and a recall of 92.8%.

Existing methods have shown good performance in controlled settings, but there remain challenges for noisy clinical data, integration of multi-view radiographs, and real-time explainability for clinical adoption. Few studies comprehensively address panoramic-to-periapical generalization or provide end-to-end frameworks that combine high-accuracy segmentation with clinically actionable explanations. This motivates the present work, which aims to use Deep Learning to perform dental radiography segmentation efficiently.

## 3. METHODOLOGY

The dataset for this study was obtained from the Kaggle repository found at (Momeni, 2023), which comprised 1269 dental radiography images in total. The train, test, and validation folders consisted of 1076, 74, and 122 images, respectively. A sample of these images is shown in Figure 1.

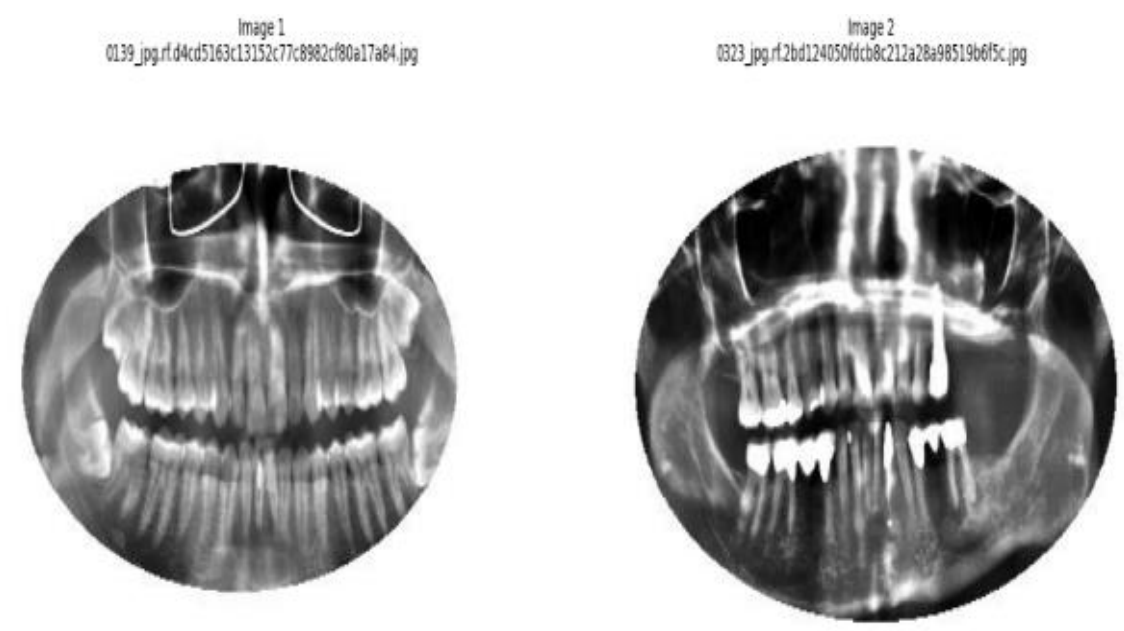

**Figure 1: Sample images from the dental radiography dataset.**

Moreover, masks for these images were generated using coordinates files that were comprised of bounding box details. We employed four segmentation models, namely, PSPNet, DeepLabV3plus, UNet, and UNet++ on these data.

### 3.1 Segmentation Models

#### 1. PSPNet

The Pyramid Scene Parsing Network, also known as PSPNet in short, was founded by (Zhao *et al.*, 2017). The architecture is accomplished through the integration of a pioneering pyramid pooling module, empowering the network to encapsulate intricate contextual details and elevate segmentation precision.

#### 2. DeepLabV3

Chen et al. (2017) found the DeepLabV3, which employ atrous convolution in cascade that handles the problem of segmenting objects at multiple scales. They have also augment their previously proposed Atrous Spatial Pyramid Pooling module, which probes convolutional features at multiple scales, with image-level features encoding global context and further boost performance (Chen *et al.*, 2017).

#### 3. UNet

The UNet architecture introduced by Ronneberger, Fischer and Brox (2015) is a convolutional neural network which is famous for image segmentation, especially in medical imaging. It has a distinct U-shape with an encoder (contracting path) and a decoder (expansive path) connected by skip connections, allowing it to capture context and precise localization for pixel-level classification, effectively mapping high-resolution inputs to high-resolution outputs (Ronneberger, Fischer and Brox, 2015).

#### 4. UNet++

UNet++ is an advanced deep learning architecture for image segmentation, building on the classic UNet by introducing nested, dense skip connections and deep supervision to bridge the semantic gap between encoder and decoder features (Zhou *et al.*, 2019). This results in more accurate segmentation, especially for varying-sized objects, by integrating multi-scale information through its complex, interconnected structure.

We selected these four representative segmentation architectures mainly due to their complementary strengths with PSPNet and DeepLabV3 offering strong multi-scale context modelling capabilities, and UNet, together with its advanced variant UNet++, demonstrating their proven effectiveness in biomedical segmentation through the use of skip connections and deep supervision. This selection allows us a balanced comparison between context-aggregation and precise localization strategies while using the same ResNet50 encoder across all models.

Alongside these four models, we tested several combinations while training by varying parameters, such as image size and batch size, along with the encoder (ResNet50). The models were trained with an Adam optimizer with a learning rate and weight decay parameter of 0.0001.

In addition, we used augmentation techniques during the training process to enhance the robustness of the results. The augmentation pipeline consisted of several metrics, including horizontal flipping (p=0.5), vertical flipping (p=0.3), random 90 degrees rotation (p=0.5), shift scale rotation (shift limit=0.1, scale limit=0.2, rotate limit =15, p=0.5), random brightness contrast adjustments (brightness limit=0.3, contrast limit=0.3, p=0.7), Gaussian noise injection (var limit=(10.0, 50.0), p=0.5), grid distortion (p=0.3), and elastic transformation (alpha = 1, sigma =50, alpha_affine = 50, p=0.3).

In addition, input resolutions were carefully selected to align with the architectural characteristics of each model family while enabling a fair comparison of segmentation performance. Specifically, we chose image sizes as multiples (or near-multiples) of 32, corresponding to the effective down-sampling stride of the ResNet50 encoder used across all models. Accordingly, PSPNet and DeepLabV3 were trained using image sizes of 256, 304, and 384, while UNet and UNet++ were trained using input sizes of 224, 256, and 300.

Moreover, batch sizes of 4, 8, and 16 were systematically evaluated to investigate the effect of batch size on training stability, generalization performance, and segmentation accuracy. These values correspond to the maximum feasible batch sizes that fit within the 16 GB GPU memory available on the Kaggle platform for the selected input resolutions.

### 3.2 Evaluation Metrics

To test the efficiency and select the best model, we mainly used three popular metrics: DICE, IoU, and accuracy.

**1. DICE**

The Dice coefficient (or Sørensen–Dice index) compares two sets (like pixel masks) to see how similar they are. After a model segments an image, the Dice score assesses how well the predicted mask matches the known correct mask. It can be calculated as in the following Equation (1).

$$DICE = 2|A\cap B|/(|A|+|B|) \quad (1),$$

where $|A\cap B|$ is the overlap (intersection) and $|A|,|B|$ are the sizes of the predicted and ground truth masks, respectively.

**2. Intersection over Union (IoU)**

IoU is a crucial metric for assessing segmentation models, commonly referred to as Jaccard's Index, as it quantifies how well the model can distinguish objects from their backgrounds in an image. IoU is a popular metric to measure localization accuracy and compute localization errors in object detection models (Rezatofighi *et al.*, 2019). It calculates the amount of overlap between two bounding boxes: a predicted bounding box and a ground truth bounding box [16], as depicted in Equation (2).

$$IoU = |A\cap B|/(|A\cup B|) = (|I|)/(|U|) \quad (2),$$

Where A and B are the prediction and ground truth bounding boxes.

**3. Accuracy**

Accuracy is also used as an additional evaluation metric in this study to measure the proportion of correctly classified pixels. It can be calculated as shown in Equation (3).

$$\text{Accuracy} = \text{Number of correctly classified pixels}/\text{Total number of pixels} \quad (3).$$

## 4. RESULTS AND DISCUSSION

In this study, we used different image sizes, including 256, 304, and 384 for PSPNet and DeepLabV3, and 224, 256, and 300 for UNet and UNet++ models, for the training purposes. Tables 1, 2, and 3 present the quantitative segmentation performance of four deep learning architectures, evaluated on the dental radiography dataset under varying image resolutions and batch sizes. We assessed the performance mainly by the DICE value, IoU value and pixel accuracy. In addition, we also considered the training and validation losses.

Among encoder-decoder models, UNet demonstrated strong performance stability across different image resolutions, with peak performance observed at 256×256 resolution, achieving an IoU of 0.6336 and a Dice score of 0.7575 using a batch size of 8. Increasing the image resolution beyond this point resulted in marginal gains or slight degradation, suggesting diminishing returns due to increased spatial redundancy.

On the other hand, the advantage of higher image resolution was more obvious in the case of PSPNet and DeepLabV3, specially at 384×384, with the IoU values of 0.6157 and 0.6338 , respectively.

This improvement highlights the effectiveness of pyramid pooling and atrous convolution in capturing multi-scale contextual information in higher-resolution dental radiographs. Pixel accuracy remained consistently high (>97%) across all models and combinations. However, given the strong class imbalance between background and dental regions, accuracy alone was found to be insufficient for reliable performance assessment, reinforcing the importance of IoU and Dice metrics for semantic segmentation.

**Table 1: Models trained with a batch size of 16, for 50 epochs**

| Model | Image Size | Metrics | | | | |
|---|---|---|---|---|---|---|
| | | Train Loss | Validation Loss | IoU | Dice | Accuracy |
| PSPNet | 256 | 0.3572 | 0.3086 | 0.6006 | 0.7283 | 0.9778 |
| | 304 | 0.3496 | 0.2835 | 0.6248 | 0.7469 | 0.9803 |
| | 384 | 0.3614 | 0.2868 | 0.6157 | 0.7384 | 0.9797 |
| Deeplabv3 | 256 | 0.3019 | 0.3180 | 0.5992 | 0.7259 | 0.9761 |
| | 304 | 0.2839 | 0.2843 | 0.6366 | 0.7601 | 0.9790 |
| | 384 | 0.2758 | 0.2838 | 0.6338 | 0.7570 | 0.9790 |
| UNet | 224 | 0.2918 | 0.2990 | 0.6265 | 0.7843 | 0.9787 |
| | 256 | 0.2791 | 0.2754 | 0.6336 | 0.7575 | 0.9802 |
| | 300 | 0.2504 | 0.2559 | 0.6587 | 0.7721 | 0.9818 |
| UNet++ | 224 | 0.2830 | 0.296 | 0.6148 | 0.7342 | 0.9789 |
| | 256 | 0.2560 | 0.2749 | 0.6331 | 0.7517 | 0.9810 |
| | 300 | 0.2850 | 0.2761 | 0.6318 | 0.7506 | 0.9802 |

Therefore, we obtained the validation IoU values for all the four models as depicted in Figures 2 and 3 to evaluate and compare their segmentation performance on the validation dataset.

According to Figure 2, DeepLabv3 generally performed better or similarly in most cases. Further, DeepLabv3 benefited more from a larger batch size (ex: 384×384) while the PSPNet model shows more variation with the input size.

When considering Figure 3, UNet++ clearly shows comparatively higher IoU scores in most configurations (ex: at a 300×300 image size with a batch size of 16). The difference is more noticeable with larger batch sizes. UNet++ has also appeared more stable across different input sizes.

**Table 2: Models trained with a batch size of 8, for 50 epochs**

| Model | Image Size | Metrics | | | | |
|---|---|---|---|---|---|---|
| | | Train Loss | Validation Loss | IoU | Dice | Accuracy |
| PSPNet | 256 | 0.3766 | 0.3087 | 0.5875 | 0.7147 | 0.9773 |
| | 304 | 0.3807 | 0.2998 | 0.5996 | 0.7248 | 0.977 |
| | 384 | 0.3834 | 0.2916 | 0.5996 | 0.7241 | 0.979 |
| Deeplabv3 | 256 | 0.2914 | 0.3035 | 0.6071 | 0.7324 | 0.9782 |
| | 304 | 0.2972 | 0.3013 | 0.6015 | 0.7276 | 0.9786 |
| | 384 | 0.2541 | 0.2743 | 0.6397 | 0.7567 | 0.9803 |
| UNet | 224 | 0.3128 | 0.3123 | 0.5988 | 0.7234 | 0.976 |
| | 256 | 0.3012 | 0.293 | 0.6165 | 0.7396 | 0.9788 |
| | 300 | 0.2649 | 0.2815 | 0.6232 | 0.744 | 0.98 |
| UNet++ | 224 | 0.268 | 0.2979 | 0.615 | 0.7377 | 0.979 |
| | 256 | 0.2593 | 0.2693 | 0.6493 | 0.7691 | 0.9804 |
| | 300 | 0.2564 | 0.2511 | 0.6719 | 0.7869 | 0.9821 |

In addition, we obtained the validation IoU heatmap as shown in Figure 4. According to Figure 4, the best results have been obtained when the data were trained with UNet++, using an image size of 300 and a batch count of 16.

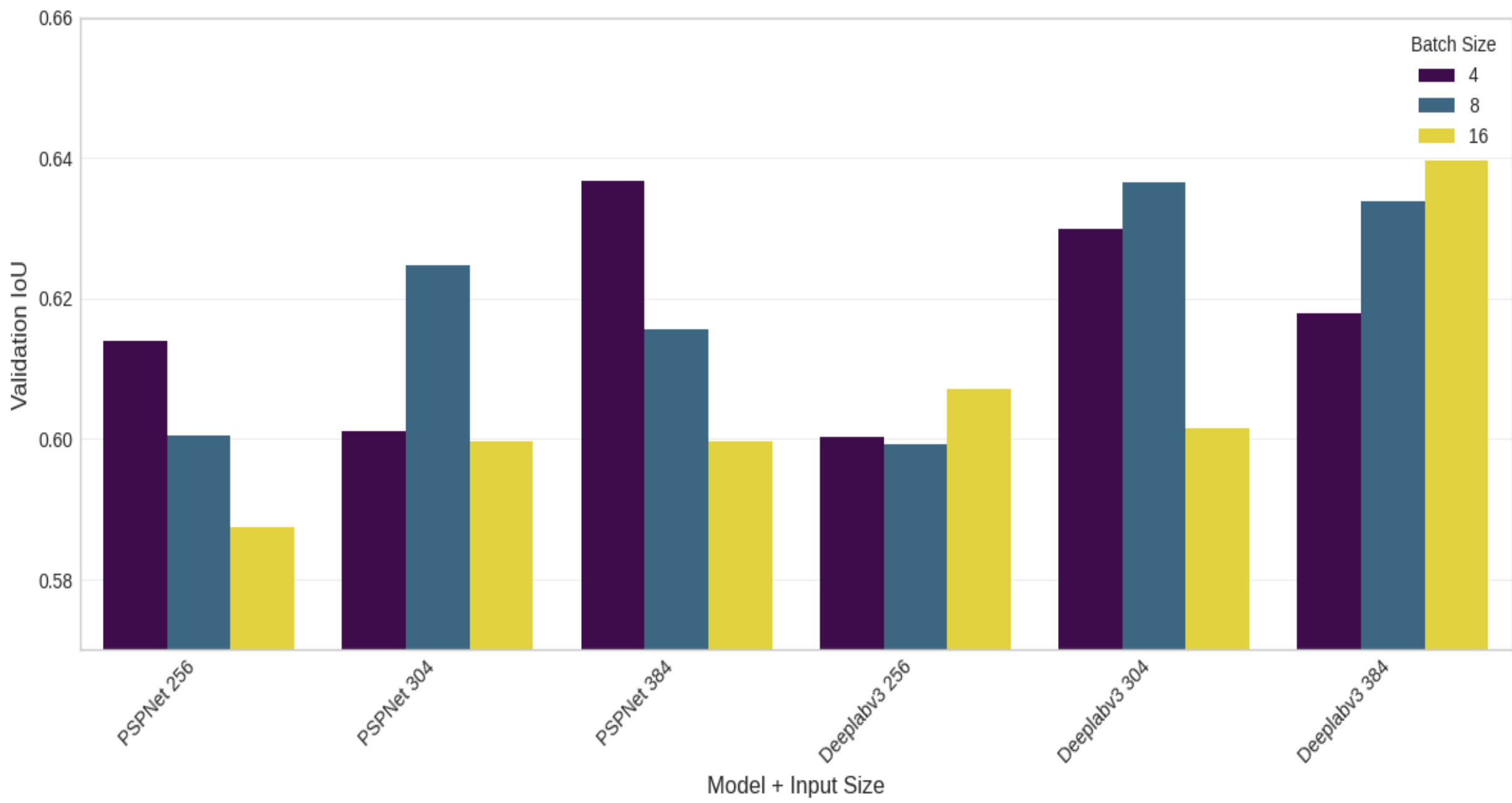

**Figure 2: PSPNet and DeepLabV3 Validation IoU Comparison.**

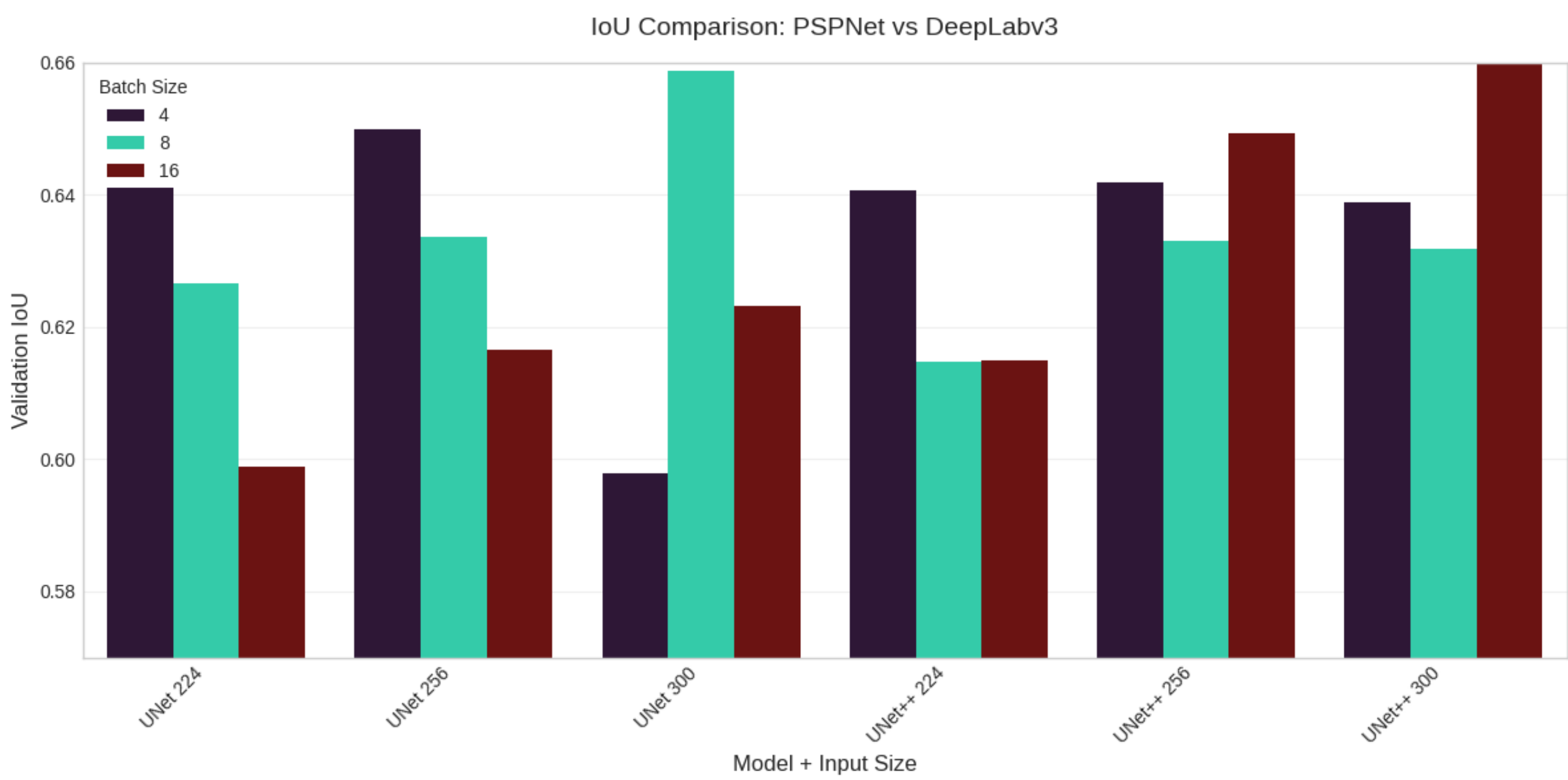

**Figure 3: UNet and UNett++ Validation IoU Comparison.**

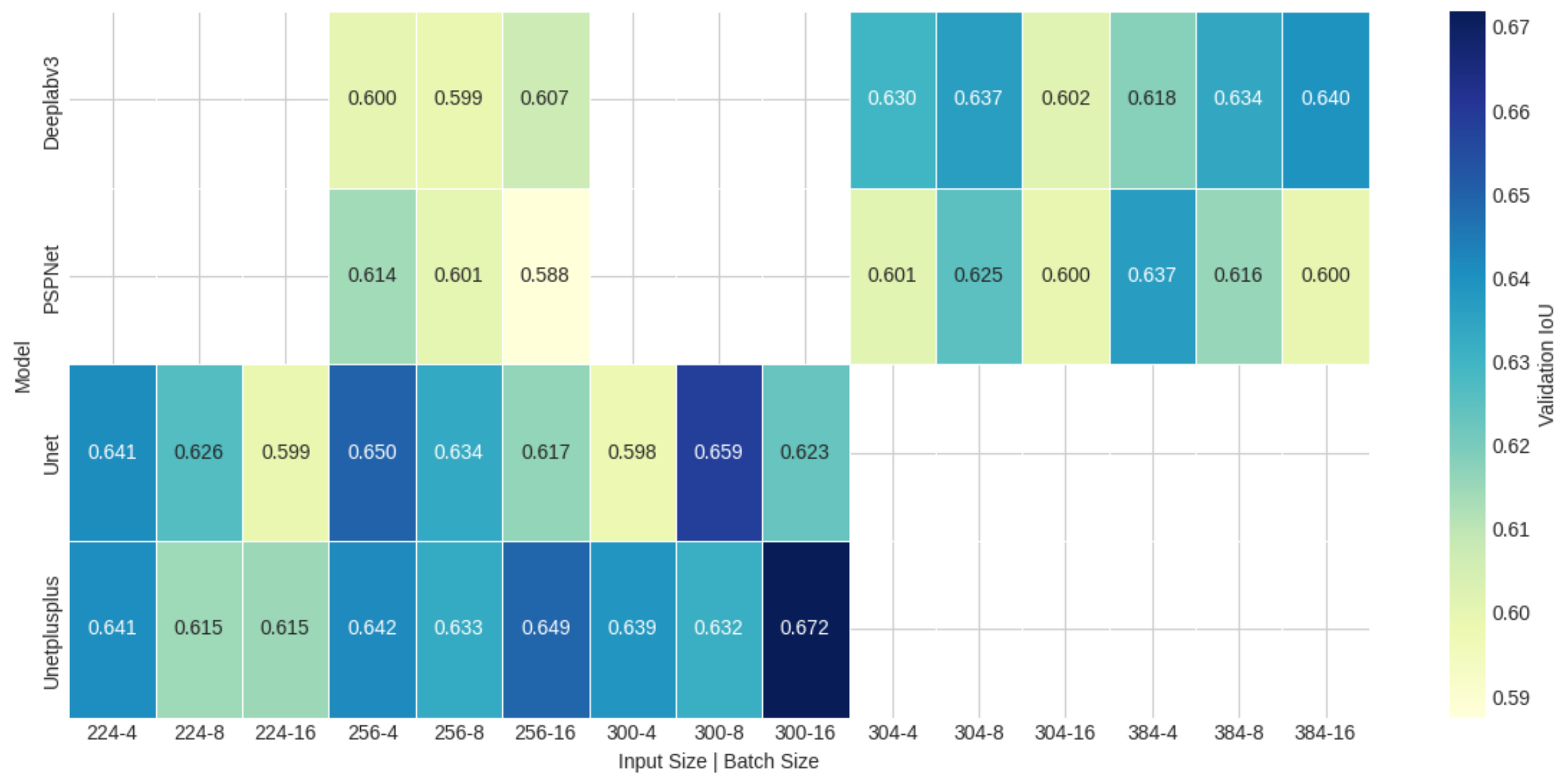

**Figure 4: Validation IoU Heatmap.**

**Table 3: Models trained with a batch size of 4, for 50 epochs**

| Model | Image Size | Metrics | | | | |
|---|---|---|---|---|---|---|
| | | Train Loss | Validation Loss | IoU | Dice | Accuracy |
| PSPNet | 256 | 0.3504 | 0.2961 | 0.6140 | 0.7402 | 0.9793 |
| | 304 | 0.3708 | 0.3206 | 0.6011 | 0.7269 | 0.9787 |
| | 384 | 0.3583 | 0.2948 | 0.6368 | 0.7637 | 0.9788 |
| Deeplabv3 | 256 | 0.3265 | 0.3390 | 0.6003 | 0.7282 | 0.9751 |
| | 304 | 0.2945 | 0.2964 | 0.6299 | 0.7535 | 0.9787 |
| | 384 | 0.2958 | 0.3016 | 0.6179 | 0.7443 | 0.9784 |
| UNet | 224 | 0.3254 | 0.2794 | 0.6411 | 0.7649 | 0.9806 |
| | 256 | 0.2648 | 0.2641 | 0.6500 | 0.7717 | 0.9812 |
| | 300 | 0.3070 | 0.3232 | 0.5978 | 0.7247 | 0.9769 |
| UNet++ | 224 | 0.2800 | 0.2857 | 0.6406 | 0.7564 | 0.9801 |
| | 256 | 0.3038 | 0.2875 | 0.6418 | 0.7633 | 0.9794 |
| | 300 | 0.2812 | 0.2798 | 0.6389 | 0.7574 | 0.9804 |

According to Figure 4, the best results have been obtained when the data were trained with UNet++, using an image size of 300 and a batch count of 16. The lowest IoU score, 0.588, was obtained for the PSPNet model, which was also comprised of a batch size of 16 with an image size of 256.

### 4.1 Effect of Image Resolution

We can observe that the image resolution has a measurable impact on segmentation quality across all models. Lower resolutions (ex, 224×224 or 256×256) enabled faster convergence but often resulted in reduced boundary precision, particularly for fine-grained dental structures such as roots and interproximal regions.

Furthermore, higher resolutions (300×300 and 384×384) improved spatial detail representation, leading to increased IoU and Dice scores for context-aware architectures such as PSPNet and DeepLabV3. However, excessively larger input sizes introduced higher computational cost and increased susceptibility to overfitting, especially when batch sizes were constrained by GPU memory.

### 4.2 Effect of Batch Size

In particular, batch size played a critical role in optimization stability and generalization. Smaller batch sizes, for example, 4, resulted in noisier gradient updates and less consistent validation performance. Conversely, batch sizes of 8 and 16 consistently produced higher IoU and Dice scores, particularly for UNet and UNet++ architectures.

Notably, UNet++ demonstrated improved performance at larger batch sizes, likely due to its dense skip connections facilitating more stable gradient propagation during training. This trend was less pronounced in PSPNet, indicating architectural sensitivity to batch size variations.

These results suggest that moderate-to-high resolutions (256–300) offer the best trade-off between segmentation accuracy and computational efficiency for dental radiography tasks. Therefore, UNet++ was selected as the best-performing model for dental radiography image segmentation, particularly with an image size of 300×300 and a batch size of 16. To prove this further, we obtained the training and validation loss curves for the best-performing model, which is shown in Figure 5.

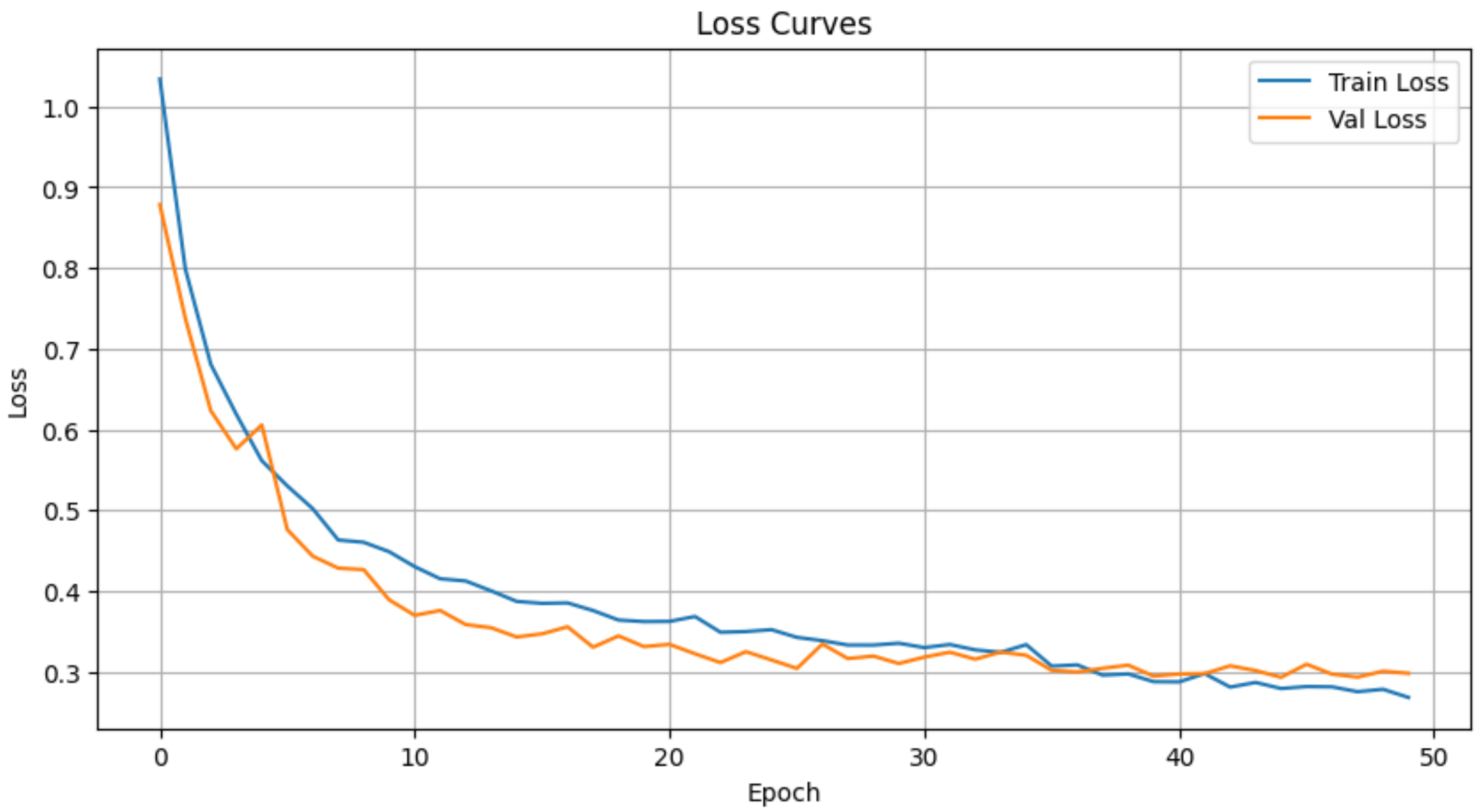

**Figure 5: Training and validation loss curves for the best model.**

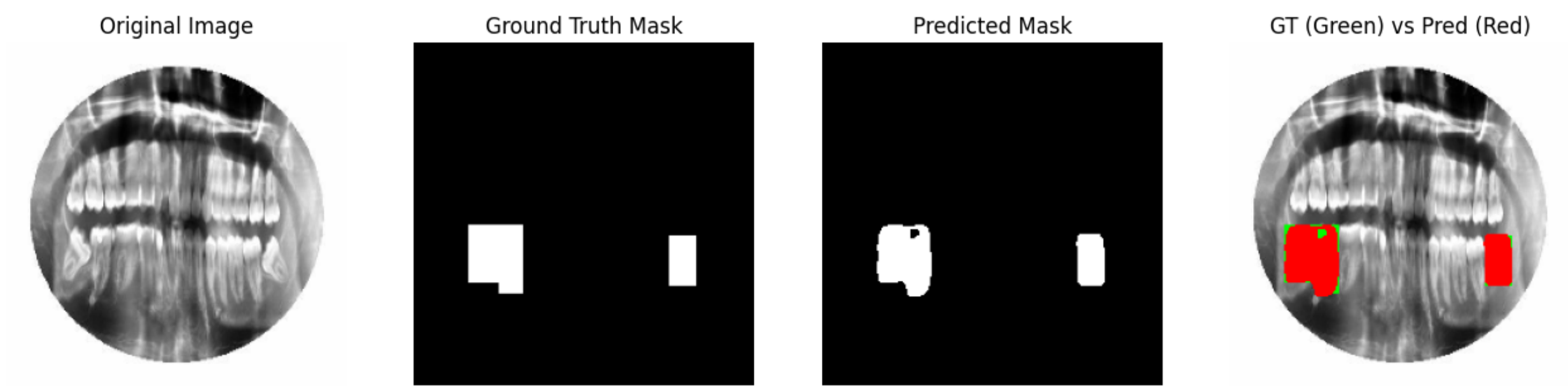

**Figure 6: Prediction performed on the best model.**

The resulting validation loss curve, which is illustrated in Figure 5, depicts that the model has not been subjected to overfitting, verifying its suitability and a sample of resulting predictions generated using the UNet++ model is shown in Figure 6.

From a clinical perspective, accurate segmentation of dental structures is essential for applications such as automated diagnosis, treatment planning, and longitudinal monitoring. Our results indicate that UNet++ based architectures provide a reliable and computationally efficient solution, making them suitable for real-time or resource-constrained clinical environments. While PSPNet and DeepLabV3 demonstrated competitive performance at higher resolutions, their increased computational overhead may limit practical deployment in routine dental workflows.

# 5. CONCLUSION

This study evaluated and compared the performance of four state-of-the-art deep learning architectures: PSPNet, DeepLabV3, UNet and UNet++ in dental radiography image segmentation. The experimentation with different image sizes and batch sizes revealed that UNet++ yields the best segmentation accuracy, specifically when using an image size of 300×300 and a batch size of 16, achieving a maximum IoU score of 0.6719. The results highlight the critical importance of hyperparameter tuning in optimizing segmentation models for dental imaging. Dental practitioners can improve diagnostic accuracy, optimize clinical workflows and deliver better patient outcomes by replacing manual techniques with fully automated, high-precision systems that enable more reliable treatment planning and monitoring.

However, this study relies on a single publicly available dataset and may have limited generalizability. Future studies will validate these models on larger multi-institutional datasets with diverse imaging modalities like cone-beam computed tomography (CBCT) and intraoral scans. Explainable AI (XAI) techniques can also be incorporated to provide interpretable segmentation insights, possibly boosting clinical trust and adoption. In conclusion, the use of such automated systems in clinical practice has a great potential to develop digital dentistry and personalized oral healthcare.

**AI ACKNOWLEDGEMENT**

The authors declare that generative AI tools, if used, were applied solely for limited technical assistance such as language editing, formatting, or similar support during manuscript preparation. All intellectual content, analysis, interpretation of results, and conclusions were developed solely by the authors. Authors take full responsibility for the accuracy, originality, and integrity of the submitted work.